\documentclass{article}
\usepackage{ijcai21}

\usepackage{times}
\usepackage{soul}
\usepackage{url}
\usepackage[utf8]{inputenc}
\usepackage[small]{caption}
\usepackage{graphicx}
\usepackage{amsmath}
\usepackage{amsthm}
\usepackage{booktabs}
\usepackage{algorithm}
\usepackage{algorithmic}
\usepackage{bm}
\usepackage{amssymb}
\usepackage{multirow}

\title{Class-Incremental Few-Shot Object Detection}

\author{
    Pengyang Li$^1$
    \and
    Yanan Li$^2$
    \and
    Han Cui$^3$
    \And
    Donghui Wang$^1$
    \affiliations
    $^1$Zhejiang University\\
    $^2$Zhejiang Lab\\
    $^3$University of California, Los Angeles
    \emails
    pyli@zju.edu.com,
    liyn@zhejianglab.com,
    elviscuihan@g.ucla.edu,
    dhwang@zju.edu.com
}

\begin{document}

\maketitle


\begin{abstract}
Conventional detection networks usually need abundant labeled training samples, while humans can learn new concepts incrementally with just a few examples. 
This paper focuses on a more challenging but realistic class-incremental few-shot object detection problem (iFSD). 
It aims to incrementally transfer the model for novel objects from only a few annotated samples without catastrophically forgetting the previously learned ones. 
To tackle this problem, we propose a novel method \emph{LEAST}, which can transfer with \emph{\textbf{L}ess forgetting, f\textbf{E}wer training resources, \textbf{A}nd \textbf{S}tronger \textbf{T}ransfer capability}. 
Specifically, we first present the transfer strategy to reduce unnecessary weight adaptation and improve the transfer capability for iFSD. 
On this basis, we then integrate the knowledge distillation technique using a less resource-consuming approach to alleviate forgetting and propose a novel clustering-based exemplar selection process to preserve more discriminative features previously learned.
Being a generic and effective method, \emph{LEAST} can largely improve the iFSD performance on various benchmarks.  
\end{abstract}

\section{Introduction}

Object detection has achieved significant improvements in both speed and accuracy based on the deep Convolutional Neural Network (CNN) \cite{ren2015faster,lin2017feature,lin2017focal,liu2018path}, but they are facing new practical challenges.
A notable bottleneck is their heavy dependency on the large training set that contains carefully annotated images. 
However, on the one hand, it is hard to collect a large and sufficiently annotated dataset that covers all the required categories for most real-world problems. On the other hand, novel classes may be continually encountered after the learning stage, e.g. detecting new living species. Training a particular model whenever these novel classes emerge is infeasible. 

Inspired by the human's remarkable ability to incrementally learn novel concepts with just a few samples, \emph{class-incremental few-shot object detection} (iFSD) is beginning to raise research attention. 
Assuming there is a detector that is well pre-trained on \emph{base} classes, iFSD aims to transfer it for \emph{novel} classes that are sequentially observed with very few training examples while not forgetting the old ones.

\begin{figure}
	\begin{center}
		\includegraphics[width=\linewidth]{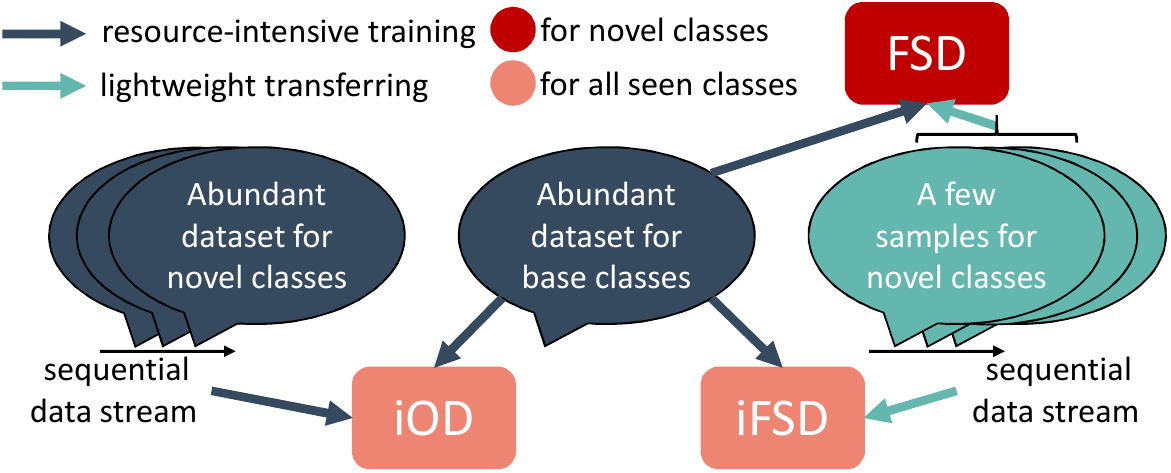}
	\end{center}
	\caption{Illustration of similarity and difference among few-shot detection (FSD), class-incremental object detection (iOD), and class-incremental few-shot detection (iFSD). iFSD uses lightweight transferring to incrementally detect novel objects from a sequential data stream, in which novel classes are offered a few samples.}
	\label{fig:diff}
\end{figure}

The majority of existing works that transfer a detection model to novel classes focus on non-incremental few-shot object detection (FSD) \cite{karlinsky2019repmet,kang2019few,wang2020frustratingly,xiao2020few} and class-incremental object detection (iOD) \cite{shmelkov2017incremental,hao2019end}. 
Fig.\ref{fig:diff} illustrates similarity and difference among FSD, iOD and iFSD. 
Compared with FSD that mainly cares for detecting novel classes while ignoring base ones, iFSD needs to attack the \emph{catastrophic forgetting} phenomenon \cite{mccloskey1989catastrophic}. 
It refers to that a neural network forgets previous knowledge when learning a new task and often happens when we simply apply FSD solutions to iFSD.  
In contrast with iOD that transfers the detector utilizing abundant labeled samples of novel classes, iFSD is more realistic and challenging since people are only willing to annotate very few samples. 
Even if we have abundant novel samples, large-scale training usually needs intensive computing resources to support, such as GPU servers or clusters. 
How to incrementally learn novel detectors under limited training resources poses another challenge in iFSD.

A straightforward idea to iFSD is integrating standard detection frameworks with class-incremental few-shot classifiers \cite{ren2019incremental,tao2020few,liu2020incremental} who use new techniques to avoid forgetting based on the insight from distillation \cite{hinton2015distilling}, i.e. previous knowledge can be retained by not perturbing the pre-trained discriminative distribution. However, due to the complicated nature of detection tasks, we need to identify multiple objects from millions of candidate regions in one single image. The above classifiers and detection networks cannot be simply merged. The very recent work \cite{perez2020incremental} proposes a class-specific weight generator to register novel classes incrementally. In each incremental novel task, it requires only a single forward pass of novel samples and does not access base classes. Although it can reduce the consumption of training resources for each novel task, it struggles to remember the knowledge learned in previous tasks and has low transfer capability to detect novel objects.

To attack aforementioned problems, we propose a novel iFSD method that incrementally detects novel objects with \emph{\textbf{L}ess forgetting, f\textbf{E}wer training resources, \textbf{A}nd \textbf{S}tronger \textbf{T}ransfer capability} (LEAST). 
It is generic and straightforward while effectively alleviating the catastrophic forgetting and economizing the consumption of training resources. 
The contributions of this paper are summarized as follows:
\begin{itemize}
	\item We first give a careful analysis of current methods that can solve the iFSD problem, and then propose a new transfer strategy that decouples class-sensitive object feature extractor from the whole detector in order to obtain stronger transfer capability with less unnecessary weight adaptation. 
	\item We integrate the knowledge distillation technique using a less resource-consuming approach in order to alleviate forgetting the previously learned knowledge. 
	\item We propose a clustering-based exemplar selection algorithm, expected to representatively capture the distribution and intra-class variance of base classes leveraging a few exemplars. 
	\item We conduct extensive experiments to demonstrate that our proposed LEAST can significantly outperform the state-of-the-arts in different settings.
\end{itemize} 

\section{Related Work}

\paragraph{Few-shot object detection.}
Modern machine learning models mainly focus on learning from abundant labeled instances. In contrast, humans can learn new concepts quickly with just a few samples, which gives rise to the recent research of few-shot learning. Most of the existing works are developed in the context of classification, which cannot be directly applied to object detection. Some previous works \cite{karlinsky2019repmet,wang2019meta,kang2019few,fan2020few} have made useful attempts in integrating few-shot classifiers with detection frameworks in order to improve the performance of novel classes. However, due to the unconstrained nature, a single test image may contain both novel and base classes. Model's knowledge retention on base classes should also be evaluated at the same time. To meet this requirement, \cite{yan2019meta,xiao2020few,wang2020frustratingly} randomly select a few base samples and then construct a balanced training set of both base and novel classes for the transfer learning stage. Thus, overfitting on novel classes and forgetting on base ones can be simultaneously alleviated to some extent. \cite{perez2020incremental} further extends this problem to the incremental learning setting, where novel tasks containing several novel classes come sequentially, and performance on all classes observed so far needs to be evaluated. 

\emph{Rehearsal}, i.e. incremental learning not only with the novel data but also with earlier data, is actually a feasible strategy to overcome \emph{catastrophic forgetting}. Under the limitation of memory usage and computational requirement, we cannot use all data in previous tasks but some exemplars instead \cite{rebuffi2017icarl}. 
The methods \cite{yan2019meta,xiao2020few,wang2020frustratingly} using randomly selected exemplars can naturally be regarded as exemplar-based solutions for iFSD.
In contrast with random selection, we propose to select a few representative exemplars with a clustering-based approach, with the hope of capturing the distribution and intra-class variance of base classes.

\paragraph{Class-incremental object detection.}
Humans can continuously learn new knowledge as their experience goes, without catastrophically forgetting previously learned knowledge. However, deep neural networks will forget what has been learned when they are trained on a new task, which is a key challenge in incremental learning. \cite{li2016learning,rebuffi2017icarl} try to attack this challenge in the context of visual classification. 
While \cite{shmelkov2017incremental,hao2019end} focus their attention on the object detection scenario, where a detector is learned from sequentially arrived data that contain disjoint objects. 
In the incremental learning stage, the knowledge distillation \cite{hinton2015distilling} technique is adopted, where extra memory for storing the frozen copy of the pre-trained model is needed. Different from using abundant data in each novel task above,  \cite{perez2020incremental} incrementally detects novel objects with just a few samples, which is more challenging and realistic.

Comparing with the above works, we attempt to integrate the knowledge distillation technique with iFSD in a generic and effective framework. 
It does not need a large amount of extra memory for the frozen copy during training, but only a little for the pre-computed logits of a few instances instead. 
Besides, LEAST can be established with either one-stage or two-stage detection frameworks. 
For a fair comparison with previous models \cite{yan2019meta,wang2020frustratingly,fan2020few,xiao2020few}, we adopt Faster-RCNN \cite{ren2015faster} as the basic architecture in this paper.

\begin{figure*}
	\begin{center}
		\includegraphics[width=\linewidth]{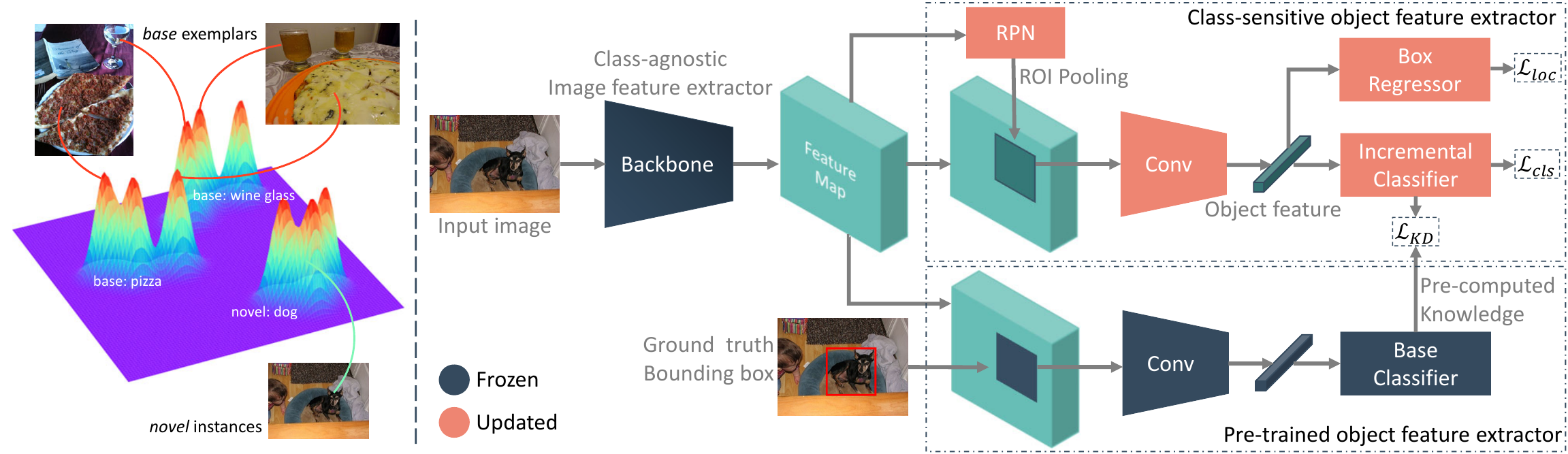}
	\end{center}
	\caption{Illustration of our proposed method based on Faster R-CNN framework. Left: our clustering-based exemplar selection that is expected to have the potential to capture modes of each base class. These exemplars and few novel instances form a balanced set for the incremental transfer stage. Right: the overall architecture. The decoupled detector and distilled knowledge are used to enhance the model's transfer capability to novel classes, without forgetting the base ones at the same time.}
	\label{fig:arch}
\end{figure*}

\section{Methodology}

\subsection{Problem Definition}

Let $C = C_b \cup C_n$ denotes the whole set of object categories. $C_b$ is the set of base classes that have a large number of training instances, annotated with object categories and bounding boxes. $C_n$ is the disjoint set of \emph{novel} classes that have only $K$ (usually less than 10) instances per class (i.e. $K$-shot detection). iFSD aims to learn a detector that can incrementally detect novel objects using $K$-shot per class. 
It may encounter a different number of novel classes in practice.
Here we consider two different settings as in \cite{perez2020incremental}. The \emph{typical} setting for iFSD is that the novel classes are added at once with a single model transfer.
In the more challenging \emph{continual} iFSD setting, the novel classes are added one by one with $|C_n|$ times model transfer.

A general iFSD solution consists of two stages: \textbf{\emph{(1)}} Pre-train stage: pre-train a standard detector on base classes; \textbf{\emph{(2)}} Incremental transfer stage: transfer the pre-trained detector to novel classes without forgetting the old ones. Considering the computational requirement and the memory limit, it should be computationally-efficient without revisiting the whole base class data. Our main focus in this paper is the essential incremental transfer stage.


\subsection{Reducing Unnecessary Weight Adaptation}
We start with a detailed analysis of current methods that can solve the iFSD problem (including exemplar-based methods mentioned in Section 2) and then propose our transfer strategy. Previous approaches can be divided into two subgroups, according to their transfer strategy in the incremental transfer stage: 
\textbf{\emph{(1)}} Fix the pre-trained detector and only adapt the last layer to novel classes (denoted as \textbf{FIX\_ALL}) \cite{wang2020frustratingly,perez2020incremental}.
\textbf{\emph{(2)}} Adapt the whole detector to novel classes (denoted by \textbf{FIT\_ALL}) \cite{yan2019meta,xiao2020few}. 
The former consumes minimal resources in transferring for novel classes while has limited generalization ability since the feature extractor is fixed. 
The latter usually has good performance on novel classes but forgets old ones since the whole network is biased towards novel objects. 
It also needs more training resources than the former methods. 
The difference in resource consumption between these two types of approaches is straightforward, and the performance difference can be found in Tab.\ref{tab:comparison_ifsd}, where TFA is the state-of-the-art method of \textbf{FIX\_ALL} and FSDetView is the state-of-the-art method of \textbf{FIT\_ALL}.

The low performance of \textbf{FIT\_ALL} on base classes make sense due to unnecessary weight adaptation.
From the architecture's perspective, we know that even if a neuron of the front layers changes a little, the final output may vary a lot after the feed-forward pass of a neural network. 
If we transfer the whole detector using limited supervision (i.e. K-shot), the discriminative distribution learned from previous classes will be further influenced as the updated layers get deeper. 
Besides, the front layers in a deep learning model usually learn generic features for an image and are well trained in the abundant training examples.
Adaptation on them is unnecessary and may cause overfitting on the few samples.

Based on the above consideration, we propose to separate the whole detector into \emph{class-agnostic image feature extractor} (unchanged during the incremental learning stage) and \emph{class-sensitive object feature extractor} (CSE) (optimized during the incremental learning stage), as is shown in Fig.\ref{fig:arch}. Usually, a deep backbone in the detection network, e.g. ResNet \cite{he2016deep}, extracts generic features for an input image and can be regarded as a class-agnostic image feature extractor. If we update the backbone with a few novel samples, its feature extraction capability will be damaged instead, possibly resulting in the forgetting. While the object feature extractor (e.g. RPN and ROI head for a two-stage detector like Faster-RCNN \cite{ren2015faster}, or FPN and extra subnets for a one-stage detector like RetinaNet \cite{lin2017focal}) is more sensitive to object categories and often extract object-specific features. 
It expects to be updated to learn more discriminative information of novel classes; otherwise, the detector will be hard to generalize to novel classes.
Thus, we propose only to optimize CSE while keep others fixed in the incremental transfer stage (denoted as \textbf{FIT\_CSE}). 

The proposed transfer strategy combines the advantages of both \textbf{FIX\_ALL} and \textbf{FIT\_ALL}. Without the unnecessary weight adaptation on class-agnostic image feature extractor, it consumes fewer training resources and is more generalized to novel objects utilizing the well learned image features.

\subsection{Less Forgetting with Knowledge Distillation}

In the incremental transfer stage, a naive method is to fine-tune the network with standard classification loss:
\begin{equation}
\mathcal{L}_{cls} = -\mathbb{E}_{x \sim \mathcal{X}}[\log p(y=y^*|x,\Phi)]
\end{equation}
where $\mathcal{X}$ denotes all candidate regions in the few available images. $y^*$ is the ground-truth label for $x$. 
$p(y|x,\Phi)$ is the classification probability for observed classes including incrementally added ones, with parameters $\Phi$.
However, since base classes are not available in the incremental transfer stage, the model tends to forget the previous knowledge catastrophically.   
Even if we use the exemplar set to store a few old samples, the model will also forget some discriminative information for previous classes due to the limited supervision.

To attack this problem, we propose to use knowledge distillation \cite{hinton2015distilling} in iFSD, inspired by its success in incremental classification \cite{li2016learning}. Although previous works \cite{shmelkov2017incremental,hao2019end} have tried integrating distillation in object detection, they still need an exact copy of the pre-trained detector to compute the learned knowledge in the discriminative distribution $p(y'|x,\Phi_{old})$. Here $y'$ belongs to the previous classes $C_{old}$. $\Phi_{old}$ denotes the parameters learned in previous tasks. In this way, much more computing resources are used than directly training the detector. In contrast, we apply knowledge distillation on positive candidate regions with pre-computed knowledge, shown in Fig.\ref{fig:arch}. 
To be specific, the previous knowledge $p(y'|x,\Phi_{old})$ for positive candidate regions $x \sim \mathcal{X}_p$ will be pre-computed through ground-truth bounding-boxes $b_{gt}$ and ROI pooling \cite{ren2015faster}.
Here $\mathcal{X}_p$ denote the set of positive candidate regions, whose Intersection over Union (IoU) with ground truth is above $\alpha$, i.e. $IoU(x, b_{gt}) > \alpha$.
Similarly, it can also be pre-computed in the one-stage detector according to the anchor owning the maximum IOU with $b_{gt}$. 
Then, it is straightforward to avoid forgetting the pre-trained knowledge:
\begin{equation}
\arg\min_{\Phi} \mathbb{E}_{x \sim \mathcal{X}_p}[KL(p(y'|x, \Phi_{old}) \| p(y'|x, \Phi))]
\label{eq:kd}
\end{equation}
where Kullback-Leibler (KL) divergence measures the forgotten information of the new discriminative distribution for $C_{old}$ with respect to the pre-trained one. 
Considering $KL(p_{old} \| p ) = H(p_{old},  p )-H(p_{old})$ and the Cross Entropy $H(p_{old})$ is irrelevant to $\Phi$, Eq.\ref{eq:kd} is equivalent to: 
\begin{align}
\resizebox{.91\linewidth}{!}{$
\displaystyle
\mathcal{L}_{kd} = \mathbb{E}_{x \sim \mathcal{X}_p} \big[\sum_{i \in C_{old}}-p(y'=i|x, \Phi_{old}) \log p(y'=i|x, \Phi)\big]
$}
\end{align}
where the classification probability for distillation is produced by scaled softmax: $p(y'=i|.) = e^{z_i/T}/\sum_j e^{z_j/T}$, and $z_i$ is the output logit of class $i$. 
$T$ is the temperature, which is suggested $T>1$ to encourage the network to better encode previously learned class similarities \cite{hinton2015distilling}. 
The final loss used in iFSD becomes: 
\begin{equation}
\mathcal{L} = \mathcal{L}_{rpn}+\mathcal{L}_{loc} + \mathcal{L}_{cls} + T^2 \mathcal{L}_{kd}
\end{equation}
where $\mathcal{L}_{rpn}$ and $\mathcal{L}_{loc}$ are the same loss as in Faster R-CNN. $T^2$ is to balance the relative contribution of $\mathcal{L}_{kd}$. 

\begin{algorithm}[tb]
	\caption{Clustering-based exemplar selection}
	\label{alg:exemplar_kmeans}
	\textbf{Input}: base dataset $\mathcal{D}_b=\{x_i,b_i,y_i\}_{i=1}^N$, base classes $C_{b}$\\
	\textbf{Require}: pre-trained feature extractor $\mathcal{F}$: $(\mathcal{X},\mathcal{B}) \rightarrow \mathbb{R}^{d}$\\
	function $\mathcal{K}$: $\mathbb{R}^{n*d} \rightarrow \mathbb{R}^{K*d}$ returns $K$ centroids of $n$ features \\
	function $\mathcal{T}$ returns the set of distinct elements for a list\\
	subscript $c$ indexes the pair $(c,value)$ and returns $value$ \\
	\textbf{Output}: exemplars $\mathcal{E}$
	\begin{algorithmic} 
		\FOR{$i \leftarrow 1,...,N$}
		\STATE $f_{i} \leftarrow \{(c, \frac{1}{|\{j|y_{ij}=c\}|}\sum_{y_{ij}=c}\mathcal{F}(x_i, b_{ij}))|c \in \mathcal{T}(y_i)\}$
		\ENDFOR
		\STATE $F \leftarrow \{f_1,...,f_{N}\}$ // features set of each image 
		
		\STATE // cluster centroids for each class
		\STATE $P \leftarrow \{(c,\mathcal{K}(\{f_{ic}|i \in \{1,...,N\} \wedge c \in y_i\}))|c \in C_b\}$
			
		\FOR{$i \leftarrow 1,...,N$}
		\STATE $q_{i} \leftarrow \{(c,\underset{k \in \{1,...,K\}}{argmin} \|f_{i,c}-P_{c,k}\|)|c \in \mathcal{T}(y_i)\}$
		\ENDFOR
		\STATE $Q \leftarrow \{q_1,...,q_{N}\}$ // cluster assignments
		
		\STATE $\mathcal{E} \leftarrow \emptyset$; $R \leftarrow \emptyset$
		\WHILE{$|R| \ne |C_{b}|*K$}
		\STATE $i \leftarrow \underset{j \in \{1,...,N\} \wedge q_j \cap R = \emptyset}{argmin} \frac{1}{|\mathcal{T}(y_j)|}\sum_{c \in \mathcal{T}(y_j)}\| f_{j,c} - P_{c,q_{j,c}} \|$ 
		\STATE $R \leftarrow R \cup q_i$ 
		\STATE $\hspace{0.5mm} \mathcal{E} \leftarrow \mathcal{E} \cup \{x_i\}$ 
		\ENDWHILE
	\end{algorithmic}
\end{algorithm}

\subsection{More Preserving with A Few Exemplars}

In order to not forget the old learned knowledge, another feasible method in the incremental transfer stage is to store a few exemplars drawn from the old training set. As discussed in Section 2, several current methods can be regarded as exemplar-based solutions for iFSD, where exemplars are selected randomly. 
However, the randomly selected exemplar set is unstable and can not guarantee to well represent the non-uniform data distributions of different classes. 
Another class-average based exemplar selection \cite{rebuffi2017icarl} that aims to approximate the class mean vector is also not suitable for iFSD because of the complex scenarios and intra-class variance in the detection task \cite{karlinsky2019repmet}. 

As is shown in the left of Fig.\ref{fig:arch}, the data distribution of each class may have multiple modes. 
To better preserve the discriminative features learned on base classes, we expect the selected exemplars could have the potential to represent these modes as many as possible. Randomly selected exemplars may not be representative, and class-average based exemplars can only capture one mode. To this problem, we propose a novel clustering-based examplar selection algorithm as follows. Details can be found in Algorithm \ref{alg:exemplar_kmeans}. 

Each image may contain multiple instances from different classes.
Thus, we need to calculate multiple features for an image.
Firstly, for simplicity, each distinct category contained in an image is represented by the averaged features of instances with the same category,
since the same category's instances in a single image are probably similar.
Secondly, for each $c \in C_b$, we use the k-means algorithm to cluster the features from images containing $c$ into $K$ clusters. 
$K$ is assumed as the number of shots in order to construct a balanced few-shot dataset between base and novel classes.
We will then obtain $K$ centroids for each category.
Finally, we progressively select the images that best approximates these learned centroids.
Since a single image may cover clusters of different classes, we hope that using at most $|C_b|*K$ images to representatively capture the discriminative features of base classes. 
Moreover, although we use the simple but effective clustering method (k-means) in this paper, other exemplar learning methods are worth trying in the future \cite{bautista2016cliquecnn,mairal2008supervised}.   

\section{Experiments}

\subsection{Experimental Setup}

\paragraph{Dataset.}
We use two popular and challenging datasets for non-incremental FSD and incremental FSD (iFSD) in this paper to evaluate the detection performance, i.e. Pascal VOC \cite{everingham2015pascal}, and MS-COCO \cite{lin2014microsoft}.
We use the same data splits as in the previous work \cite{kang2019few,wang2019meta,yan2019meta,wang2020frustratingly} for FSD and the work \cite{perez2020incremental} for iFSD, respectively. 
In MS-COCO, there are 80 classes in total, which include the whole 20 classes in Pascal VOC.  The 60 categories disjoint with Pascal VOC are used as \emph{base} classes, while the remaining 20 categories are used as \emph{novel} ones. Each novel class has $K \in \{1, 5, 10\}$ samples, and 10 random sample groups are considered in this paper, following \cite{wang2020frustratingly}. 
For the MS-COCO dataset, we use 5000 images as in \cite{kang2019few} from the validation set for evaluation, and the rest images containing at least one base instance in train/val sets for pre-training.
For the Pascal VOC dataset, we use the 2007 test set for testing.

\paragraph{Evaluation metric.}

To evaluate the detection performance, we use the average precision (AP) with IOU threshold from 0.5 to 0.95 of the top 100 detections and the corresponding average recall (AR) as the evaluation metrics. For non-incremental FSD, only the performance on novel classes is tested. While for iFSD, the performance on both base and novel classes needs to be tested.
To evaluate the model's comprehensive performance between base and novel domains in a balanced way, we also report their harmonic mean value, i.e. $HM(x, y) = 2xy /(x+y)$, same as another incremental few-shot scenario \cite{cermelli2020few}.
Some works only report the mean score of all classes, which ignores the significance of novel classes.
Since the number of novel classes is usually much smaller than base classes, the simple mean score will bias a lot to base classes and then can not well evaluate the comprehensive performance.

\paragraph{Implementation details.}
We use Faster-RCNN as our basic detection architecture and ResNet-101 as the backbone following \cite{wang2020frustratingly}. We train the model using the SGD optimizer with a momentum of 0.9 and a weight decay of 0.0001. The parameters $\alpha$ and $T$ are set to $0.7$ and $20$ respectively. During the pre-training stage, we train the standard Faster-RCNN on base classes for 6 epochs with a learning rate of 0.01, which is decreased by 10 after 4 epochs.  We freeze the class-agnostic image feature extractor during the incremental transfer stage and then train our proposed method for 10 epochs with a learning rate of 0.001. 
Our code is uploaded as an attachment.

\begin{table}
	\centering
	\begin{tabular}{c|c|cc}
		\toprule
		Transfer & & \multicolumn{2}{c}{Novel Classes} \\
		Strategy & Method & AP & AR \\
		\midrule
		\multirow{2}*{FIX\_ALL} 
		& MetaDet* & 7.1 & 15.5 \\
		& TFA* & 9.1 & - \\
		\midrule
		\multirow{3}*{FIT\_ALL} 
		& Meta-RCNN* & 8.7 & 17.9 \\
		& Attn-RPN & 11.1 & - \\
		& FSDetView* & 12.5 & 25.7 \\
		\midrule
		FIT\_CSE & LEAST* & \textbf{12.8} & \textbf{27.5} \\
		\bottomrule
	\end{tabular}
	\caption{Non-incremental few-shot object detection performance on COCO with 10 shots. ‘-’: Not reported results. ‘*’: Results averaged over multiple random runs.}
	\label{tab:comparison_nifsd}
\end{table}

\subsection{Non-Incremental Few-Shot Detection}

When all novel classes are added at once in one incremental learning stage, and only novel classes are focused, the iFSD problem naturally degenerates into the vanilla FSD problem. In this case, our method (LEAST) can be regarded as an effective solution to the non-incremental FSD, along with the specific advantage of not forgetting base classes at the same time. We compare LEAST with several state-of-the-arts on MS-COCO in Tab.\ref{tab:comparison_nifsd}, which are MetaDet \cite{wang2019meta}, Meta-RCNN \cite{yan2019meta}, TFA \cite{wang2020frustratingly}, Attn-RPN \cite{fan2020few}, and FSDetView \cite{xiao2020few}.
From Tab.\ref{tab:comparison_nifsd}, we can observe that LEAST can achieve comparable results on novel classes with non-incremental FSD approaches. AP and AR are even a little higher than the state-of-the-art FSDetView.  Since unknown objects in a test image may cover all possible categories, the class similarities and discriminative information previously learned will also benefit novel classes' performance. It validates the effectiveness of LEAST. Furthermore, comparing with all these competitors that generally improve the detection performance of novel classes while sacrificing the performance on base ones, LEAST has a significant advantage of not forgetting previous knowledge. 


\begin{figure*}
\begin{minipage}{0.67\linewidth}
\centering
\setlength{\tabcolsep}{5pt}{
	\begin{tabular}{c|c|c|cc|cc|cc}
		\toprule
		& & & \multicolumn{2}{c|}{Base Classes} & \multicolumn{2}{c|}{Novel Classes} & \multicolumn{2}{c}{Harmonic Mean}\\
		Shots & Method & EB & AP & AR & AP & AR & AP & AR \\
		\midrule
		\multirow{4}{0.1\linewidth}{1} 
		& ONCE & & 17.9 & 19.5 & 0.7 & 6.3 & 1.3 & 9.5 \\
		& TFA* & $\surd$ & \textbf{31.9} & - & 1.9 & - & 3.6 & - \\
		& LEAST-NE* & & 24.6 & \underline{35.8} & \textbf{4.4} & \textbf{21.6} & \textbf{7.5} & \textbf{26.9} \\
		& LEAST* & $\surd$ & \underline{29.5} & \textbf{43.1} & \underline{4.2} & \underline{18.0} & \underline{7.4} & \underline{25.4} \\
		\midrule
		\multirow{4}{0.1\linewidth}{5} 
		& ONCE & & 17.9 & 19.5 & 1.0 & 7.4 & 1.9 & 10.7 \\
		& TFA* & $\surd$ & \textbf{32.3} & - & 7.0 & - & 11.5 & - \\
		& LEAST-NE* & & 25.2 & \underline{36.4} & \textbf{9.4} & \textbf{27.6} & \underline{13.7} & \underline{31.4} \\
		& LEAST* & $\surd$ & \underline{31.3} & \textbf{44.5} & \underline{9.3} & \underline{24.4} & \textbf{14.3} & \textbf{31.5} \\
		\midrule
		\multirow{5}{0.1\linewidth}{10} 
		& ONCE & & 17.9 & 19.5 & 1.2 & 7.6 & 2.2 & 10.9 \\
		& TFA* & $\surd$ & \textbf{32.4} & - & 9.1 & - & 14.2 & - \\
		& FSDetView* & $\surd$ & 10.5 & - & \underline{12.5} & 25.7 & 11.4 & - \\
		& LEAST-NE* & & 23.1 & \underline{34.2} & \underline{12.5} & \textbf{30.3} & \underline{16.2} & \underline{32.1} \\
		& LEAST* & $\surd$ & \underline{31.3} & \textbf{44.3} & \textbf{12.8} & \underline{27.5} & \textbf{18.2} & \textbf{33.9} \\
		\bottomrule
	\end{tabular}
}
\captionof{table}{Typical iFSD performance on COCO. ‘EB’: Exemplar-based methods for iFSD. ‘-’: Not reported results. ‘*’: Results averaged over multiple random runs. \textbf{Bold} and \underline{underline} indicate the best and the second-best.}
\label{tab:comparison_ifsd}
\end{minipage}
\hfill
\begin{minipage}{0.32\linewidth}
	\centering
	\begin{minipage}{0.8\linewidth}
		\includegraphics[width=\linewidth]{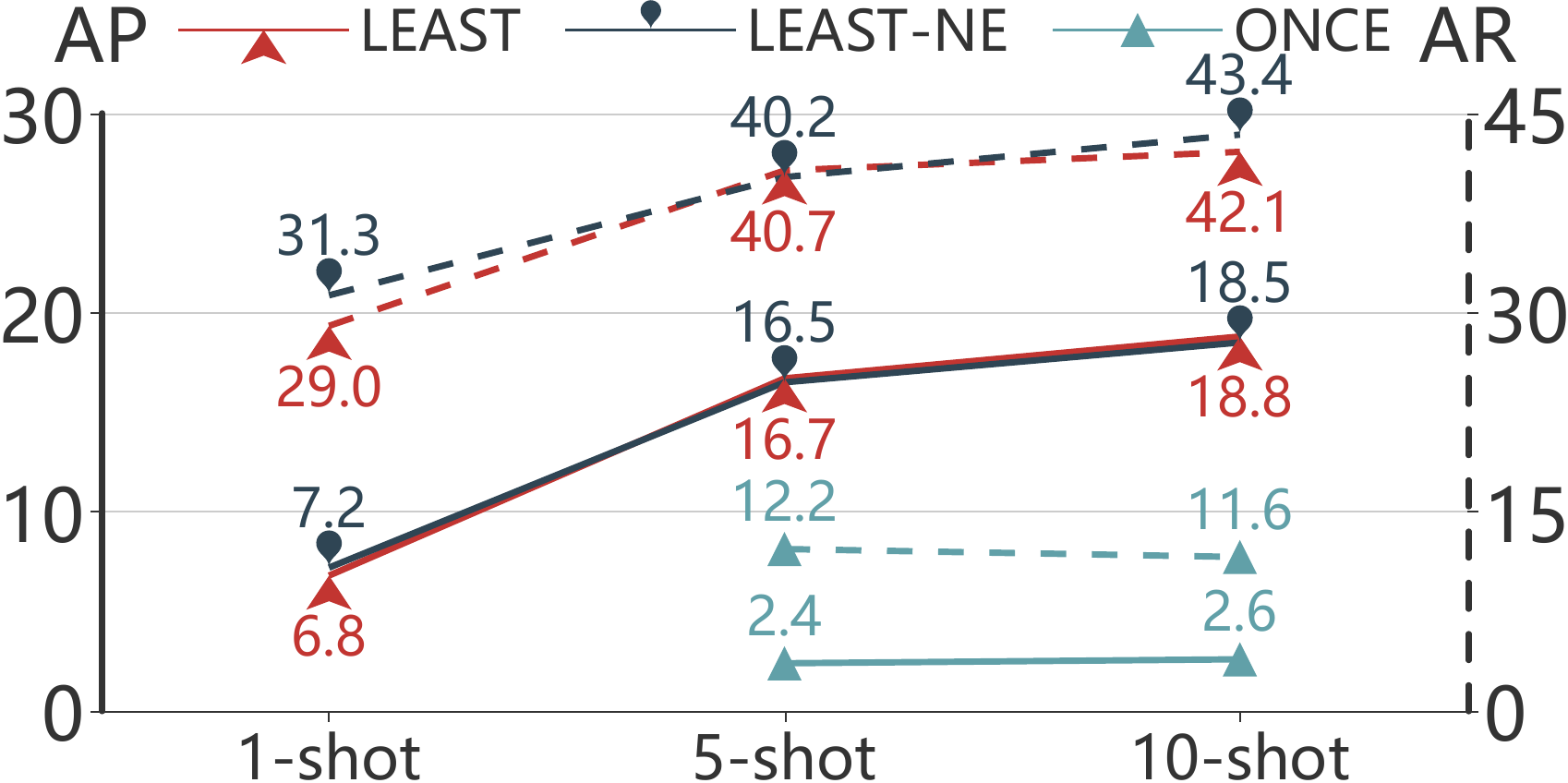}
	\end{minipage}
	\vfill
	\begin{minipage}{0.8\linewidth}
		\includegraphics[width=\linewidth]{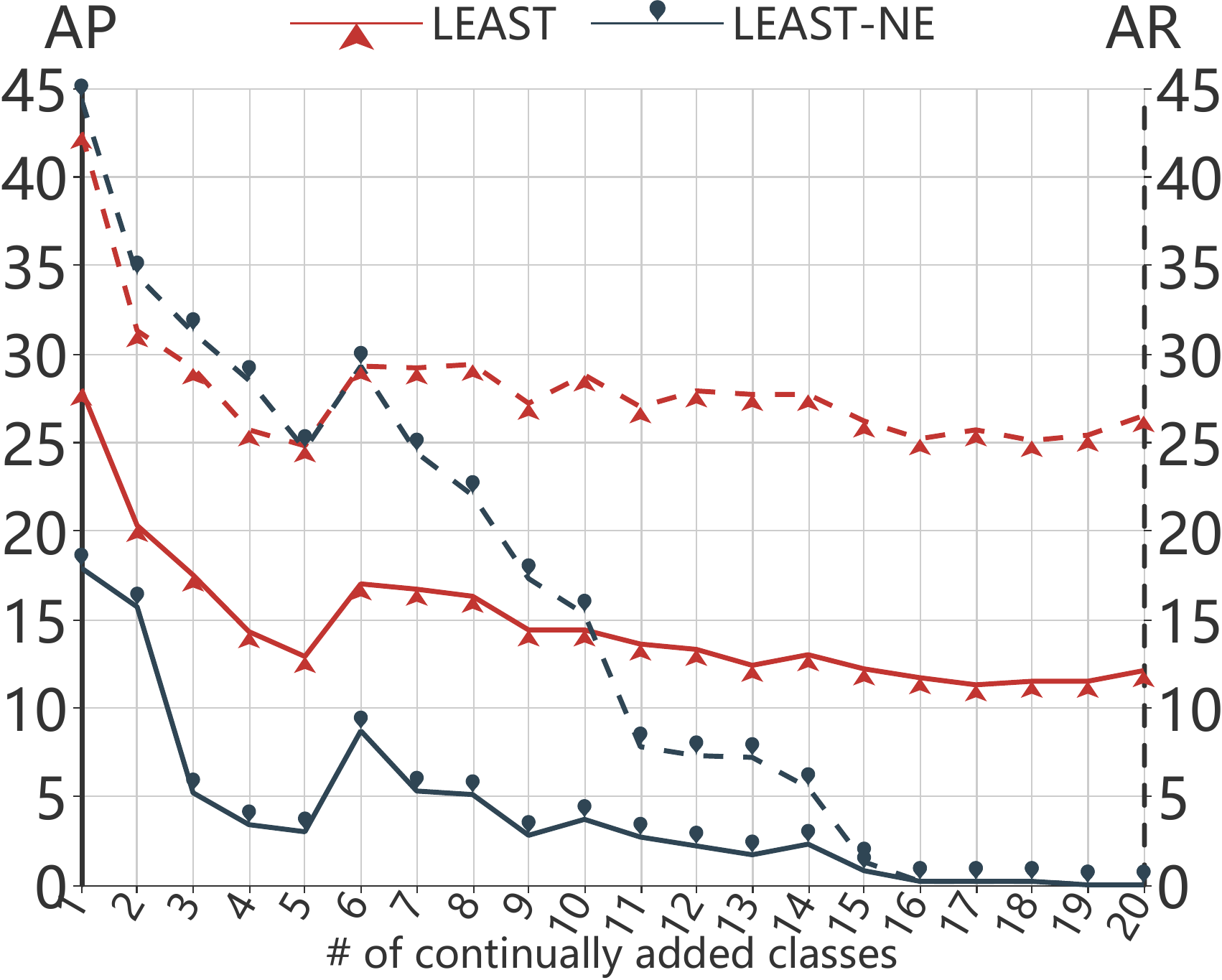}
	\end{minipage}
	\caption{Top: cross-domain evaluation from COCO to VOC using 10 shots. Bottom: continual iFSD performance on COCO with 10 shots. Dashed lines are for AR. Solid lines are for AP.}
	\label{fig:voc_cifsd}
\end{minipage}
\end{figure*}

\subsection{Class-Incremental Few-Shot Detection}


\paragraph{Typical iFSD results. }
We first evaluate the iFSD performance under the \emph{typical} setting on MS-COCO, where all the novel classes are added at once with one incremental transfer session. In this case, as mentioned in Section 2, some methods for FSD  can be naturally regarded as exemplar-based solutions for iFSD. We compare LEAST with several state-of-the-arts in Tab.\ref{tab:comparison_ifsd}: ONCE \cite{perez2020incremental}, TFA \cite{wang2020frustratingly} and FSDetView \cite{xiao2020few}. We also report our proposed method's performance using no exemplars for a fair comparison with methods without exemplars, denoted as `LEAST-NE'. 

From Tab.\ref{tab:comparison_ifsd}, we have the following observations: (1) LEAST performs better on novel classes than all competitors, with a second highest performance on base ones.
TFA freezes the whole feature extractor so that it obtains the best base AP that is a little higher than ours, but its performance on novel classes is explicitly limited.
Compared with these methods that have a large performance gap between base and novel classes, LEAST can better balance these two domains and thus have a much higher HM value. Even without exemplars, ours can still achieve promising results on avoiding forgetting and adapting to novel classes. This can validate the effectiveness of using knowledge distillation and transferring with less unnecessary weight adaptation. (2) Using a few exemplars selected by our proposed approach is generally beneficial to iFSD. It can largely improve the performance on previous tasks, and then the HM value, which indicates that previous discriminative features are better preserved.

\paragraph{Continual iFSD results. }
We then evaluate the iFSD performance under the \emph{continual} setting, where the novel classes are added one at a time with $|C_n|$ model updates. We report the harmonic mean performance of base classes and all novel classes added so far in the bottom of Fig.\ref{fig:voc_cifsd}. 
We can see that as novel classes are processed sequentially, the performance of LEAST first decreases and then levels off. While without exemplars for base classes, LEAST-NE decreases to 0 after 15 incremental transfer stages. This means that the previously learned knowledge is catastrophically forgotten. Comparing with \emph{typical} iFSD performance (i.e. AP 18.2 and AR 33.9), the final performance (i.e. AP 12.1 and AR 26.5) of \emph{continual} iFSD after 20 sessions is lower. It is reasonable since fewer model updates naturally forget less previous knowledge.  
As ONCE does no report the harmonic mean and has no released code, it does not appear in Fig.\ref{fig:voc_cifsd}.
Yet we can find its performance on the \emph{typical} setting (i.e. AP 2.2 and AR 10.9) is lower than ours on the \emph{continual} setting.
As the \emph{continual} setting is more challenging, we can legitimately infer that LEAST consistently outperforms it.

\paragraph{Cross-domain iFSD evaluation.}
We also evaluate the \emph{typical} iFSD performance in a cross-domain setting from MS-COCO to Pascal VOC. 
The performance is only evaluated on novel classes since VOC contains no objects of base classes. 
As shown at the top of Fig.\ref{fig:voc_cifsd}, our method (either with or without the selected exemplar set) significantly outperforms the previous competitors in terms of both AR and AP, which verifies the efficacy of LEAST in the cross-domain setting. Comparing with the results of MS-COCO in Tab.\ref{tab:comparison_ifsd}, the performance on VOC is higher on both AP and AR. The performance gap is reasonable since MS-COCO images contain more complex scenarios from both base and novel objects.


\begin{table}
	\centering
	\setlength{\tabcolsep}{3.5pt}{
		\begin{tabular}{l|cc|cc|cc}
			\toprule
			& \multicolumn{2}{c|}{Base Classes} & \multicolumn{2}{c|}{Novel Classes} & \multicolumn{2}{c}{HM} \\
			Method & AP & AR & AP & AR & AP & AR \\
			\midrule
			FIT\_ALL+$d$+$e$ & 27.6 & 39.7 & 11.7 & 25.9 & 16.4 & 31.3 \\
			FIX\_ALL+$d$+$e$ & \textbf{32.7} & \textbf{45.9} & 9.3 & 23.8 & 14.5 & 31.3 \\
			\midrule
			FIT\_CSE & 8.3 & 15.3 & 10.1 & \underline{28.4} & 9.1 & 19.9 \\
			FIT\_CSE+$e$ & 29.7 & 43.8 & 10.9 & 25.5 & 15.9 & 32.2 \\
			FIT\_CSE+$d$ & 23.1 & 34.2 & 12.5 & \textbf{30.3} & 16.2 & 32.1 \\
			FIT\_CSE+$d$+$e_{a}$ & 30.6 & 44.0 & 12.2 & 26.8 & 17.4 & 33.3 \\
			FIT\_CSE+$d$+$e_{r}$ & 30.5 & 44.0 & \underline{12.6} & 27.5 & \underline{17.8} & \underline{33.8} \\
			FIT\_CSE+$d$+$e$ & \underline{31.3} & \underline{44.3} & \textbf{12.8} & 27.5 & \textbf{18.2} & \textbf{33.9} \\
			\bottomrule
		\end{tabular}
	}
	\caption{Ablation study of each proposed component and their potential competitors on COCO with 10 shots.}
	\label{tab:ablation_component}
\end{table}

\paragraph{Ablation studies.} 
In the ablation study, we test the influence of the proposed transfer strategy, distillation loss, and the exemplar selection method for iFSD in Tab.\ref{tab:ablation_component}. 
It is clear that the proposed transfer strategy \textbf{FIT\_CSE} significantly outperforms \textbf{FIX\_ALL} and \textbf{FIT\_ALL}.
Meanwhile, the distillation loss (denoted by $\bm{d}$) largely improves novel class performance since it can preserve the previously learned discriminative information. 
Besides capable of remembering base classes, the proposed clustering-based exemplar selection algorithm ($\bm{e}$) performs better than random selection (\bm{$e_r$}) and class-average based selection ($\bm{e_a}$) \cite{rebuffi2017icarl}, when we use the same number of selected exemplars for a fair comparison.
It shows that the proposed selection method preserves more modes for base classes.

\section{Conclusion}

We delved into the realistic and challenging problem: \emph{class-incremental few-shot object detection}, which aims at incrementally detecting novel objects from just a few labeled samples while without forgetting the previously learned ones. We proposed a generic and effective method that uses relatively fewer training resources and can still have stronger transfer capability with less forgetting. Extensive experimental results under different settings verified its effectiveness.

\bibliographystyle{named}
\bibliography{bib}

\end{document}